\documentclass[11pt]{article}

\usepackage[preprint]{acl}

\usepackage{times}
\usepackage{latexsym}

\usepackage[T1]{fontenc}

\usepackage[utf8]{inputenc}

\usepackage{microtype}

\usepackage{inconsolata}

\usepackage{graphicx}

\newcommand{\eg}{\textit{e}.\textit{g}.,~}
\usepackage{booktabs}
\usepackage{multirow}
\title{A Compliance-Preserving Retrieval System for Aircraft MRO Task Search}

\author{Byungho Jo \\
  AI Convergence Reserach Center, Inha University \\ 
  \texttt{bhjo12@inha.ac.kr}}

\begin{document}
\maketitle
\begin{abstract}
Aircraft Maintenance Technicians (AMTs) spend up to 30\% of work time searching manuals—a documented efficiency bottleneck in MRO operations where every procedure must be traceable to certified sources.
We present a compliance-preserving retrieval system that adapts LLM reranking and semantic search to aviation MRO environments by operating alongside, rather than replacing, certified legacy viewers.
The system constructs revision-robust embeddings from ATA chapter hierarchies and uses vision-language parsing to structure certified content, allowing technicians to preview ranked tasks and access verified procedures in existing viewers.
Evaluation on 49k synthetic queries achieves >90\% retrieval accuracy, while bilingual controlled studies with 10 licensed AMTs demonstrate 90.9\% top-10 success rate and 95\% reduction in lookup time—from 6-15 minutes to 18 seconds per task.
These gains provide concrete evidence that semantic retrieval can operate within strict regulatory constraints and meaningfully reduce operational workload in real-world multilingual MRO workflows.
\end{abstract}

\section{Introduction}
Aircraft Maintenance Technicians (AMTs) regularly rely on certified maintenance manuals to locate the exact tasks required to inspect or repair aircraft systems~\cite{FAA_CFR_43_13a_2024}.
Despite their centrality to aviation safety, these manuals have grown into extremely large and intricate information sources—often exceeding tens of thousands of pages organized through multi-level Air Transport Association (ATA) structures~\cite{faa2012technical,aircraftcommerce2018}.
As a result, field reports~\cite{TATEM2012} indicate that up to 30\% of a AMTs’ work time is spent searching for the correct procedure. This challenge is not merely an industry inefficiency but represents a fundamental information-retrieval bottleneck—technicians must translate informal, problem-driven queries into highly structured, deeply nested documentation that was never designed for natural-language access.

\begin{figure}[t]
\centering
\small
\ttfamily
\begin{minipage}{0.95\linewidth}
\begin{tabular}{@{}l@{}}
\textcolor{gray}{Chapter 21: Air Conditioning}\\
\textbf{Chapter 22: Auto Flight} \textcolor{blue}{(120 tasks)}\\
\textcolor{gray}{\quad ... (10 chapters omitted)}\\
\textbf{Chapter 32: Landing Gear} \textcolor{blue}{(155 tasks)} \\
\quad +-- 32-09 Main Landing Gear \textcolor{blue}{(100 tasks)}\\
\quad +-- \textbf{32-41 Brake System} \textcolor{blue}{(55 tasks)}\\
\quad\quad\quad +-- 32-41-20 Brake Disconnect\\
\quad\quad\quad +-- \textcolor{gray}{... (6 components)}\\
\quad\quad\quad +-- \textbf{32-41-31 Gear Brake}\\
\quad\quad\quad\quad\quad +-- \textbf{401 Removal}\\
\quad\quad\quad\quad\quad |\quad\quad +-- \textbf{32-41-41-000-801} \textit{Removal}\\
\quad\quad\quad\quad\quad |\quad\quad +-- \textbf{32-41-41-400-801} \textit{Installation}\\
\quad\quad\quad\quad\quad +-- 601 Inspection\\
\textcolor{gray}{\quad ... (15 chapters omitted)}\\
Chapter 72: ENGINE \textcolor{blue}{(180 tasks)}\\
\end{tabular}
\end{minipage}

\caption{ATA chapter-based manual structure illustrating the hierarchical complexity that AMTs must navigate to locate specific maintenance tasks. A representative example, \texttt{Ch.32 $\rightarrow$ 32-41 $\rightarrow$ 32-41-31 $\rightarrow$ 401 $\rightarrow$ 32-41-41-000-801}, demonstrates a five-level navigation path with over fifty branching options. Numbers in \textcolor{blue}{blue} indicate the task counts at each level.}

\label{fig:ata-hierarchy}
\end{figure}
As illustrated in Figure~\ref{fig:ata-hierarchy}, certified manuals impose an additional structural burden: their ATA chapter hierarchy a deep, tree-structured index that technicians must manually navigate. Reaching a single end-task often requires traversing four to six nested levels, each containing dozens of branching options, before encountering several candidates with near-identical titles. For example, tasks such as “Brake Valve Removal” and “Brake Shuttle Valve Removal” appear across different ATA substructures with minimal lexical distinction. This combination of hierarchical depth, dense branching, and high lexical ambiguity makes keyword-based search fundamentally unreliable, frequently forcing technicians to open and compare multiple candidates before determining the correct procedure.

Prior work in AI for aircraft maintenance largely sidesteps this retrieval bottleneck. Research on predictive maintenance, and AR/VR-based training systems~\cite{Jo_Oh_Ha_Lee_Hong_Neumann_You_2014, tugce2025ar} has focused on optimizing maintenance execution rather than helping AMTs locate the correct procedure in the first place. Meanwhile, recent NLP efforts in aviation—such as safety-report generation~\cite{ray2023asrs} therefore do not engage with the rigid, hierarchical structure of certified manuals. 
Standard RAG frameworks typically present rewritten or synthesized text to the user, but MRO regulations require technicians to read the certified manual itself—even when the regenerated content is semantically identical. This regulatory constraint prevents direct adoption of standard RAG pipelines and motivates compliance-preserving retrieval approaches.
To the best of our knowledge, no existing work addresses semantic task retrieval under this unique combination of constraints: immutable documentation, deep hierarchical indexing, and AMTs-generated natural-language queries.

To address this gap, we introduce a compliance-preserving assistive retrieval system that enables natural-language task lookup without modifying OEM manuals or viewers. The key idea is to exploit the stability of ATA metadata: task titles and hierarchy paths change far less frequently than full text. 
Our system builds revision-robust task embeddings exclusively from this metadata while using a vision–language model (VLM) only to structure page-level content for previews—not for retrieval—thus avoiding any generated or altered text. At query time, an LLM re-ranks a candidate set but never sees or produces procedural content, preserving certification boundaries.

We evaluate the system through two complementary studies.
(1) A large-scale synthetic benchmark of 49k AMT-style queries tests robustness to paraphrasing, synonyms, and typos.
(2) A bilingual human study with ten licensed AMTs examines real-world performance using English and Korean queries over English-only manuals, reflecting common multilingual MRO environments.

The contributions of this study are as follows:
\begin{itemize}
\item We formalize task lookup in certified maintenance manuals as a semantic retrieval problem while respecting immutability, traceability, and revision-control constraints.

\item We demonstrate a scalable manual-to-knowledge conversion pipeline using ATA metadata and VLM-based structuring that requires only minimal post-editing.

\item Through synthetic and human evaluations, we show that the method achieves >90\% retrieval accuracy and reduces lookup time by over 95\%—from minutes to seconds, suggesting a practical path for adoption in airline MRO workflows.

\end{itemize}

\section{Related Work}
\subsection{Artificial Intelligence in Aircraft MRO}
Prior AI research in aircraft MRO has focused on operational execution (AR/VR-guided maintenance) and predictive analytics(failure forecasting) but has not addressed the fundamental bottleneck of locating correct procedures within certified manuals.
Augmented Reality systems~\cite{Jo_Oh_Ha_Lee_Hong_Neumann_You_2014, tugce2025ar} assume technicians have already identified the correct task, while predictive maintenance~\cite{yang2022predictive} forecasts component failures without addressing manual navigation complexity. The challenge of semantic task retrieval under regulatory constraints—where manuals cannot be modified and every procedure must be traceable—remains unexplored.

\subsection{Large Language Models in Aviation}
Recent studies have explored domain-adapted LLMs in aviation for Q\&A for pilot training~\cite{wang2024aviationgpt}, safety report summarization~\cite{ray2023asrs}, and traffic management~\cite{abdulhak2024chatatc}.
However, MRO task retrieval requires mapping queries to exact certified procedures with full audit trails—a regulatory constraint that prohibits LLM-generated content. To our knowledge, no prior work addresses semantic retrieval under these constraints.
\section{Proposed System Architecture}
\begin{figure*}
    \centering
    \includegraphics[width=1\linewidth]{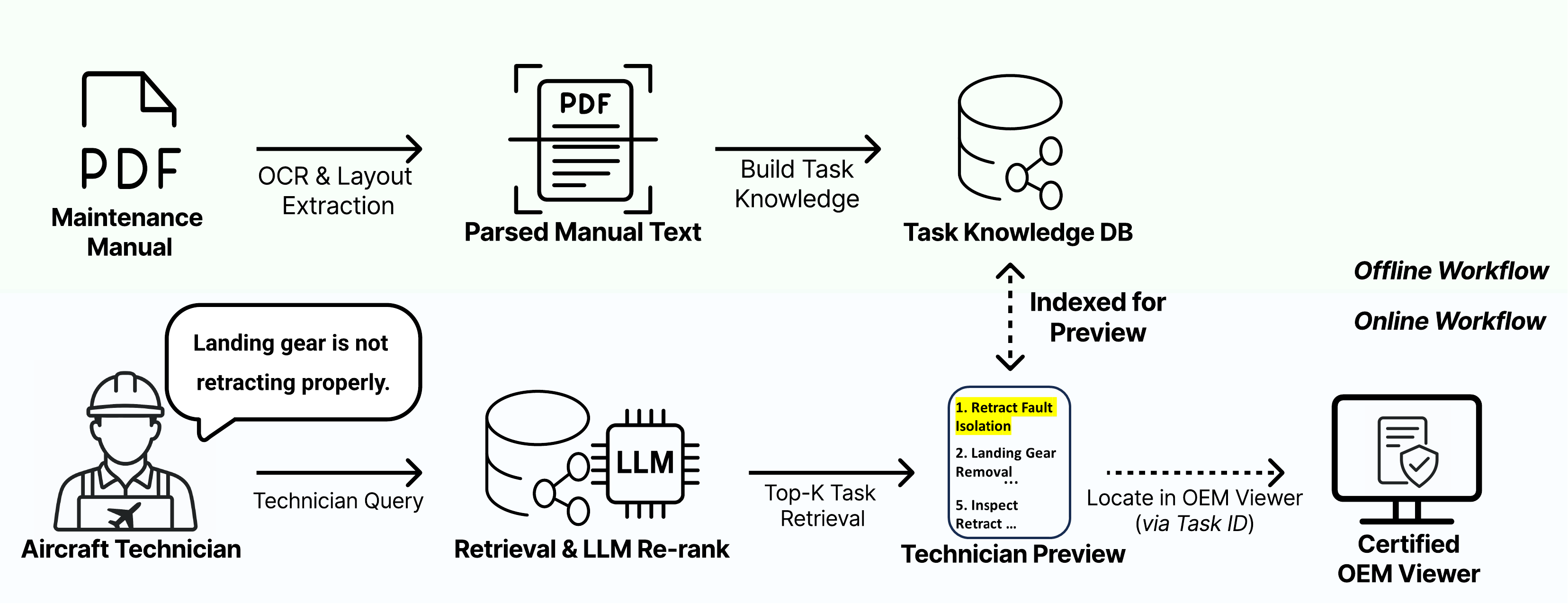}
    \caption{Offline workflow extracts and structures tasks from maintenance PDF manuals into a Task Knowledge DB. During the online workflow, a technician query triggers Top-K retrieval with LLM re-ranking, previews the ranked tasks, and opens the certified procedure in the official viewer, maintaining full compliance while reducing lookup time from minutes to seconds.}
    \label{fig:framework}
\end{figure*}

To address the dual challenge of regulatory compliance and operational efficiency, our framework introduces an assistive retrieval system that functions independently of certified OEM viewers, which cannot be modified under aviation regulations. As illustrated in Figure~\ref{fig:framework}, the system operates in two main stages: an offline knowledge structuring stage and an online retrieval stage.

\subsection{Offline Maintenance Task Knowledge Structuring}
The knowledge representation pipeline runs once per manual revision cycle and consists of two complementary processes:

\noindent\textbf{Revision-robust Task Embedding.}
To minimize re-indexing frequency across manual revisions, we construct task embeddings from stable semantic components: ATA chapter hierarchy titles concatenated with final task titles (e.g., "Landing Gear → Brake System → Gear Brake → Removal"). We exclude task procedural text (which changes frequently across revisions).

\noindent\textbf{Manual-to-Knowledge Conversion.}
We extract structured task-level representations from PDF manuals using a vision-language model that captures verbatim text and layout information. Given the manuals' clear hierarchical structure (Section → Sub-task → Step), the extracted text is transformed into structured records using rule-based parsing that preserves original identifiers and metadata, requiring only minimal post-editing. Technicians can preview these structured tasks before accessing the procedure in the certified viewer.
\subsection{Manual Retrieval Pipeline Architecture}
\label{subsec2-3}
When a technician submits a query (\eg ``Landing gear is not retracting properly''), the system retrieves the top-$N$ semantically relevant candidate tasks from the embedding database.

\noindent\textbf{LLM-assisted Re-ranking.}
To refine accuracy, the LLM receives a structured prompt containing: (1) the technician's natural-language query, (2) the top-50 candidate tasks with their ATA IDs, hierarchy paths, and titles, and (3) an instruction to output only a JSON-formatted array of re-ranked indices based on semantic relevance (e.g., [3, 15, 7, 1, ...]). This strict output format prevents content generation and hallucination—the LLM performs ranking only, without access to or ability to modify procedural content.

\noindent\textbf{Fail-safe Fallback.}
If the LLM fails to return valid JSON, the system defaults to the baseline dense retrieval rankings, ensuring robustness in safety-critical workflows.

\noindent\textbf{Structured Task Presentation.}
The Top-N re-ranked tasks are presented with their ATA IDs, titles, and metadata. This reflects real-world MRO workflows where technicians routinely review 5-10 similar tasks due to functional overlap across subsystems (e.g., "Brake Valve Removal" may exist in multiple brake assemblies), enabling them to identify the correct task before accessing the certified OEM viewer.

\section{Experiments}
Our evaluation followed a two-phase design to progressively validate the proposed system under increasing realism. 
Phase~1 established controlled baseline performance using synthetic benchmark queries, while Phase~2 validated real-world simulated utility through a human study with practicing maintenance technicians.

\subsection{Synthetic Benchmark}
\noindent\textbf{Synthetic Query Generation.}
We constructed a large-scale benchmark using publicly available Boeing 737 AMM and FIM indices~\cite{boeing737amm, boeing737FIM}, covering 8,229 tasks across major aircraft systems. Using GPT-4o~\cite{openai2024gpt4ocard}, we generated six queries per task in both full-sentence and keyword styles. To simulate field conditions, we created typo-injected variants, resulting in 49,643 evaluation queries total. Query design was informed by experienced AMTs to reflect realistic workplace search patterns.

\noindent\textbf{Evaluation Metric.} We report Hit@k (k = 1, 5)—the percentage of queries where the ground-truth task appears within the top-k results.
We focus on Hit@5 as the primary metric, reflecting operational requirements identified through AMT interviews: technicians routinely review 5-10 candidate procedures due to functional overlap between similar tasks (e.g., "Brake Assembly Removal" across multiple landing gear positions). While Hit@1 measures exact-match precision, Hit@5 captures the system's ability to deliver a manageable candidate set—the actual deployment criterion. The automated scale of synthetic evaluation allows us to measure both metrics, with Hit@5 performance >90\% indicating reliable operational utility.

\subsection{Human Study Design}
\noindent\textbf{Participants}.
We conducted a controlled study with 10 licensed aircraft maintenance technicians currently employed at a commercial airline in Korea. The participant group comprised technicians with diverse experience levels ranging from 1 to 10 years, including both junior technicians and senior experts, ensuring the generalizability of our findings across different skill levels commonly found in real-world MRO operations.

\noindent\textbf{Experimental Protocol}.
We designed a controlled retrieval evaluation using 10 AMM maintenance tasks spanning diverse ATA chapters (landing gear, fuel systems, flight controls, etc.) and action types (removal, installation, inspection, lubrication). To simulate the airline's cloud-based PDF viewer environment, we deployed a web-based interface that mirrored their operational workflow: query submission, ranked task preview, and direct PDF access for verification.

Participants were provided with official AMM task titles (e.g., "Escape Slide Pack and Cover Removal") and instructed to reformulate them into natural workplace language without directly copying. For example, a participant might query "how to remove escape slide" or "slide pack cover disassembly procedure." This tested the system's ability to bridge the semantic gap between certified documentation and technicians' everyday phrasing.

Each participant completed ten retrieval tasks twice—once in English and once in Korean—resulting in 200 searches total. This bilingual design evaluated cross-lingual performance, as the knowledge base contained only English ATA structures and task titles. A multilingual embedding model (BGE-M3~\cite{chen-etal-2024-m3}) enabled Korean queries to retrieve English task embeddings in the same semantic space, with Qwen3-8B-FP16~\cite{yang2025qwen3technicalreport} used for re-ranking.

For each query, the system presented the top-10 ranked candidate tasks with metadata (ATA ID, title, chapter). Participants clicked on candidates to open the corresponding AMM PDF pages directly in the viewer, replicating the intended deployment workflow where technicians preview ranked results before accessing certified procedures in their existing system. They verified whether the correct target task appeared within the top-10 results and recorded the outcome as Success or Failure.

The system automatically logged task completion times from query submission to final verification. Participants also reported their estimated times for locating the same manuals using conventional workplace methods and during their early-career (junior) period, enabling comparative analysis across experience levels. Details of the system implementation are provided in the Appendix.

\begin{table*}[t]
\centering
\small
\caption{Task retrieval accuracy (Hit@k) across manual types and query conditions. Hit@5 (bold) represents the primary operational metric, as technicians routinely review 5-10 candidates in practice. Hit@1 is reported for reference but understates operational utility due to functional overlap between similar tasks.}
\label{tab:consolidated_retrieval}
\begin{tabular}{l|cc|cc|cc|cc|cc}
\toprule
\multirow{2}{*}{\textbf{Model}} & \multicolumn{2}{c|}{\textbf{Overall}} & \multicolumn{2}{c|}{\textbf{AMM}} & \multicolumn{2}{c|}{\textbf{AMM-typo}} & \multicolumn{2}{c|}{\textbf{FIM}} & \multicolumn{2}{c}{\textbf{FIM-typo}} \\
& Hit@1 & Hit@5 & Hit@1 & Hit@5 & Hit@1 & Hit@5 & Hit@1 & Hit@5 & Hit@1 & Hit@5 \\
\midrule
BM25 & 46.79 & 73.06 & 54.68 & 87.57 & 32.50 & 63.38 & 66.46 & 90.26 & 49.01 & 73.63 \\
Dense Retrieval & 60.65 & 85.34 & 66.94 & 90.59 & 49.29 & 78.84 & 66.91 & 89.49 & 59.88 & 82.69 \\
\midrule
Llama3.3-70B & \textbf{79.24} & 91.64 & \textbf{79.23} & 91.75 & \textbf{76.76} & 88.96 & \textbf{78.06} & 93.05 & \textbf{82.96} & \textbf{92.89} \\
Qwen3-32B & 78.25 & \textbf{91.81} & 78.18 & \textbf{92.70} & 76.22 & \textbf{89.20} & 76.79 & \textbf{93.10} & 81.82 & 92.33 \\
Qwen3-14B & 77.65 & 91.58 & 78.42 & 92.68 & 74.86 & 88.96 & 76.44 & 92.78 & 80.87 & 92.00 \\
Phi4-14B & 77.91 & 90.38 & 77.96 & 91.04 & 76.10 & 87.93 & 76.61 & 91.44 & 80.97 & 91.21 \\
Qwen3-8B & 76.41 & 90.81 & 76.80 & 91.75 & 73.99 & 88.27 & 75.05 & 91.78 & 79.82 & 91.52 \\
Qwen3-4B & 72.58 & 91.02 & 73.25 & 92.13 & 66.49 & 88.08 & 74.17 & 92.55 & 76.61 & 91.43 \\
\bottomrule
\end{tabular}
\end{table*}
\noindent\textbf{Evaluation Metrics}. We collected: (1) Retrieval success rate—whether the target task appeared within the top-5 and top-10 results. 
(2) Task completion time (TCT, from query submission to final verification in PDF viewer). (3) Cross-lingual performance (English vs. Korean accuracy and TCT). (4) Comparative time efficiency (system TCT vs. self-reported manual lookup times for current and junior-level experience).

Unlike synthetic evaluation where Hit@1 and Hit@5 can be precisely measured, the human study prioritizes ecological validity: technicians interact with the system as they would in deployment, reviewing the top-5 and top-10 ranked list and clicking through to verify the correct procedure—mirroring real-world operational workflow.

\subsection{Performance Analysis}
\noindent\textbf{Synthetic Benchmark Performance.}
Table~\ref{tab:consolidated_retrieval} presents retrieval accuracy across 49,643 evaluation samples. The system achieves 91.64\% Hit@5 with LLM re-ranking (LlaMA3.3-70B~\cite{grattafiori2024llama3herdmodels}), a 6.3pp improvement over nomic dense retrieval~\cite{nussbaum2025nomic} (85.34\%). Critically, compact models maintain >90\% Hit@5 (Qwen3-8B: 90.81\%, Qwen3-4B: 91.02\%),confirming deployment feasibility in resource-constrained MRO environments.

The system demonstrates robust performance under noisy input conditions—a critical requirement for field operations where technicians input queries under time pressure or adverse conditions. On typo-injected queries, LLM re-ranking maintains >88\% Hit@5 across all manual types, while lexical baselines degrade significantly (BM25~\cite{BM25}: 63.38\% on AMM-typo vs. 87.57\% clean). This semantic resilience, combined with consistent >90\% Hit@5 across model sizes, ensures technicians can reliably identify correct procedures within manageable candidate sets—directly addressing the documented efficiency bottleneck.

\noindent\textbf{Human Study Results and Analysis.}
Overall, the system achieved 90.9\% top-10 success rate (179/197 queries, 95\% CI: 86.0–94.1\%) and 86.3\% top-5 success rate (170/197 queries, 95\% CI: 80.8-90.4\%), with mean Task Completion Time (TCT) of 18.0 seconds (95\% CI: 12.5–23.6s).
The 90.9\% top-10 success rate aligns with synthetic Hit@5 performance (91.64\%), validating that controlled benchmark evaluation translates to real-world operational utility.

\noindent\textbf{Cross-lingual Performance.} 
As shown in Table~\ref{tab:ir_perf}, cross-lingual performance revealed a 9.9-point gap between English (95.9\% top-10 SR) and Korean (86.0\% top-10 SR) queries, with mean TCTs of 14.2 and 22.2 seconds, respectively.
This disparity reflects both embedding limitations and the linguistic characteristics of aviation maintenance: certified terminology is standardized in English, and technicians commonly use English task names in practice. 
Several participants reported that Korean phrasing felt less natural and often reverted to English terminology when formulating precise queries. Nevertheless, the 86.0\% Korean success rate demonstrates that multilingual embeddings provide a viable path forward, even in domains where English remains the dominant operational language.
 
\noindent\textbf{Time Efficiency Gains.} 
Compared to conventional manual lookup methods, our system delivered substantial efficiency improvements. As shown in Table~\ref{tab:tct}, traditional lookup required an average of 6.35 minutes for experienced AMTs and 15.41 minutes for juniors. In contrast, our system reduced lookup times to approximately 18 seconds on average, corresponding to a 95.3\% reduction for experienced and 98.1\% for junior AMTs—absolute time savings of 6.1 and 15.1 minutes, respectively.

\begin{table}[t]
\centering
\small
\caption{Overall and cross-lingual retrieval performance in the human study. The knowledge base contains only English-language manuals.}
\begin{tabular}{lccc}
\toprule
 & English & Korean & Overall \\
\midrule
Top-5 SR (\%) & 88.7 & 84.0  & 86.3  \\
Top-10 SR (\%) & 95.9 & 86.0  & 90.9  \\
95\% CI & 89.9--98.4 & 77.9--91.5 & 86.0--94.1 \\
TCT (s) & 14.2 & 22.2 & 18.0 \\
\bottomrule
\label{tab:ir_perf}
\end{tabular}
\end{table}

\begin{table}[t]
\centering
\small
\caption{Time efficiency gains compared to traditional manual lookup}
\begin{tabular}{lcc}
\toprule
Metric & Experienced & Junior \\
\midrule
Traditional Method & 6.35 min & 15.41 min \\
Our System & $\sim$0.30 min & $\sim$0.30 min \\
Time Reduction & 95.3\% & 98.1\% \\
Absolute Savings & 6.1 min & 15.1 min \\
\bottomrule
\label{tab:tct}
\end{tabular}
\end{table}

\begin{table}[t]
\centering
\small
\caption{Cumulative distribution of task completion times (successful queries)}
\begin{tabular}{lc}
\toprule
Time Threshold & Success Rate (Cumulative) \\
\midrule
$\leq$ 10 seconds & 57.5\% \\
$\leq$ 20 seconds & 79.3\% \\
$\leq$ 30 seconds & 88.3\% \\
$\leq$ 60 seconds & 96.6\% \\
\bottomrule
 \label{tab:cumulative}
\end{tabular}
\end{table}

\noindent\textbf{Task Completion Time Distribution.} 
As shown in Table~\ref{tab:cumulative}, among successful queries, 57.5\% were resolved within 10 seconds, 79.3\% within 20 seconds, 88.3\% within 30 seconds, and 96.6\% within 60 seconds. 
These results indicate that the majority of lookups can be completed in real time, supporting the suitability of the framework for operational deployment in maintenance environments.

\noindent\textbf{Operational Impact.} 
The combination of high retrieval accuracy (90.9\%) and dramatic time reduction (>95\%) directly addresses the critical inefficiency identified by domain experts—namely, that manual lookup can consume up to 30\% of technician work time. 
While our evaluation measured the time required to locate a single manual entry, technicians typically perform multiple lookups during a maintenance session, so the cumulative savings scale proportionally.
These findings provide strong empirical evidence that our compliance-preserving framework delivers substantial productivity gains while safeguarding the precision and regulatory compliance required in aviation MRO operations.

\noindent\textbf{Failure Case Analysis.} 
While the framework is highly accurate, we observe two primary sources of retrieval failure. 
First, simple lexical issues (\eg synonyms, abbreviations) occasionally cause mismatches. 
Second, context-dependent ambiguity arises when tasks with nearly identical titles differ only by procedural sequence—such as cleaning procedures distinguished by maintenance stage.
Third, cross-lingual failures in Korean queries often stem from translation ambiguity: aviation-specific English terms (\eg "thrust reverser", "brake shuttle valve") lack standardized Korean equivalents, leading technicians to paraphrase inconsistently. This lexical variability degrades both embedding alignment and LLM semantic understanding, resulting in lower ranking accuracy for Korean queries compared to English.
These cases highlight the limitations of relying solely on task titles and underscore the need for context-aware and multilingually robust refinement in MRO workflows.

\subsection{Knowledge Structuring Quality}
To validate the offline knowledge structuring pipeline, we evaluated the vision-language parsing accuracy on production manuals. Using Qwen 2.5-VL-72B \cite{bai2025qwen25vltechnicalreport} with rule-based post-processing, we achieved >99\% precision/recall with <3\% character error rate across 20 A320 and 20 B737 manuals, as shown in Table~\ref{tab:ocr_results}.
These results confirm the VLM as a dependable, low-overhead front end for automated text extraction across aircraft types with minimal post-editing. This fidelity is crucial for maintaining knowledge-base integrity, the foundation for downstream retrieval and re-ranking. 
\begin{table}[t]
  \centering
  \small
   \caption{Text extraction accuracy of Qwen 2.5-VL on A320 and B737 family PDF-based aircraft maintenance manuals.}
  \begin{tabular}{lcccc}
    \toprule
    \textbf{Dataset} & \textbf{Precision}$\uparrow$ & \textbf{Recall}$\uparrow$ & \textbf{F1}$\uparrow$ & \textbf{CER}$\downarrow$ \\ \midrule
    A320-Family & 99.39 & 99.82 & 99.57 & 1.14 \\
    B737-Family & 99.64 & 99.27 & 99.45 & 2.57 \\ \midrule
    Total & 99.51 & 99.54 & 99.51 & 1.85\\ \bottomrule
  \end{tabular}
  \label{tab:ocr_results}
\end{table}

\section{Conclusion}
We present a compliance-preserving retrieval system that enables AMTs to locate maintenance tasks using natural-language queries without modifying certified systems. 
Our evaluation demonstrates over 90\% retrieval accuracy across both synthetic benchmarks (>90\% Hit@5 on 49k queries) and real-world validation (90.9\% top-10 success rate with 10 licensed AMTs in bilingual English/Korean queries), reducing lookup time by over 95\%—from 6-15 minutes to approximately 18 seconds. 
These results validate the practical utility of LLM-augmented retrieval in safety-critical MRO workflows while maintaining full regulatory compliance. Future work should incorporate procedural context to resolve ambiguities between similar task titles, and extend capabilities to multimodal queries and cross-document linking for a comprehensive MRO cognitive assistant.
\bibliography{custom}

\end{document}